\documentclass[sigconf]{acmart}
 
\usepackage{graphicx}
\usepackage{amsmath}
\usepackage{booktabs}
\usepackage{algorithm}
\usepackage{algorithmic}
\usepackage{amsfonts}
\usepackage{bbm}

\newcommand*{\origrightarrow}{}

\renewcommand*{\textrightarrow}{\fontfamily{cmr}\selectfont\origrightarrow}

\usepackage{multirow}
\usepackage{rotating}

\makeatletter
\def\Hline{%
\noalign{\ifnum0=`}\fi\hrule \@height 1pt \futurelet
\reserved@a\@xhline}
\makeatother

\usepackage{balance}

\def\BibTeX{{\rm B\kern-.05em{\sc i\kern-.025em b}\kern-.08emT\kern-.1667em\lower.7ex\hbox{E}\kern-.125emX}}
    
\copyrightyear{2019}
\acmYear{2019}
\setcopyright{acmcopyright}
\acmConference[CIKM '19] {The 28th ACM International Conference on Information and Knowledge Management}{November 3--7, 2019}{Beijing, China}
\acmBooktitle{The 28th ACM International Conference on Information and Knowledge Management (CIKM'19), November 3--7, 2019, Beijing, China}
\acmPrice{15.00}
\acmDOI{10.1145/3357384.3357890}
\acmISBN{978-1-4503-6976-3/19/11} 

\begin{document}

\fancyhead{}

\title[Learning More with Less]{Learning More with Less: Conditional PGGAN-based\\ Data Augmentation for Brain Metastases Detection Using\\ Highly-Rough Annotation on MR Images}

\author{Changhee Han$^{1,2,3,4}$}
\author{Kohei Murao$^{1,2}$}
\email{han@nlab.ci.i.u-tokyo.ac.jp}
\affiliation{%
  \institution{$^{1}$Research Center for Medical Big Data,\\  National Institute of Informatics}
  \city{Tokyo}
  \country{Japan}
}

\author{Tomoyuki Noguchi$^{2}$}
\author{Yusuke Kawata$^{2}$}
\author{Fumiya Uchiyama$^{2}$}
\affiliation{%
  \institution{$^{2}$Department of Radiology, National Center for\\ Global Health and Medicine}
  \city{Tokyo}
  \country{Japan}
}

\author{Leonardo Rundo$^{3}$}
\affiliation{%
  \institution{$^{3}$Department of Radiology,\\ University of Cambridge}
  \city{Cambridge}
  \country{UK}
  }
 
\author{Hideki Nakayama$^{4}$}
\author{Shin'ichi Satoh$^{1,4}$}
\affiliation{%
\institution{$^{4}$Graduate School of Information Science and Technology, The University of Tokyo}
  \city{Tokyo}
  \country{Japan}
  }

%
\renewcommand{\shortauthors}{Han and Murao, et al.}

%
\begin{abstract}
Accurate Computer-Assisted Diagnosis, associated with proper data wrangling, can alleviate the risk of overlooking the diagnosis in a clinical environment. Towards this, as a Data Augmentation (DA) technique, Generative Adversarial Networks (GANs) can synthesize additional training data to handle the small/fragmented medical imaging datasets collected from various scanners; those images are realistic but completely different from the original ones, filling the data lack in the real image distribution. However, we cannot easily use them to locate disease areas, considering expert physicians' expensive annotation cost. Therefore, this paper proposes Conditional Progressive Growing of GANs (CPGGANs), incorporating highly-rough bounding box conditions incrementally into PGGANs to place brain metastases at desired positions/sizes on $256 \times 256$ Magnetic Resonance (MR) images, for Convolutional Neural Network-based tumor detection; this first GAN-based medical DA using automatic bounding box annotation improves the training robustness. The results show that CPGGAN-based DA can boost $10\%$ sensitivity in diagnosis with clinically acceptable additional False Positives.
Surprisingly, further tumor realism, achieved with additional normal brain MR images for CPGGAN training, does not contribute to detection performance, while even three physicians cannot accurately distinguish them from the real ones in Visual Turing Test.
\end{abstract}

%
%


%

%

%
\maketitle

\section{Introduction}
Accurate Computer-Assisted Diagnosis (CAD) with high sensitivity can alleviate the risk of overlooking the diagnosis in a clinical environment. Specifically, Convolutional Neural Networks (CNNs) have revolutionized medical imaging, such as diabetic eye disease diagnosis~\cite{gulshan2016development}, mainly thanks to large-scale annotated training data. However, obtaining such annotated medical big data is demanding; thus, better diagnosis requires intensive Data Augmentation (DA) techniques, such as geometric/intensity transformations of original images~\cite{ronneberger2015u,milletari2016v}. Yet, those augmented images intrinsically have a similar distribution to the original ones, leading to limited performance improvement; in this context, Generative Adversarial Network (GAN)~\cite{goodfellow}-based DA can boost the performance by filling the real image distribution uncovered by the original dataset, since it generates realistic but completely new samples showing good generalization ability; GANs achieved outstanding performance in computer vision, including $21\%$ performance improvement in eye-gaze estimation~\cite{Shrivastava}.

Also in medical imaging, where the primary problem lies in small and fragmented imaging datasets from various scanners~\cite{Rundo}, GAN-based DA performs effectively: researchers improved classification by augmentation with noise-to-image GANs (e.g., random noise samples to diverse pathological images)~\cite{frid2018gan} and segmentation with image-to-image GANs (e.g., a benign image with a pathology-conditioning image to a malignant one)~\cite{shin2018medical,jin2018ct}. Such applications include $256 \times 256$ brain Magnetic Resonance (MR) image generation for tumor/non-tumor classification~\cite{Han2}. Nevertheless, unlike bounding box-based object detection, simple classification cannot locate disease areas and rigorous segmentation requires physicians' expensive annotation.

\begin{figure}[t]
  \centering
  \centerline{\includegraphics[width=\columnwidth]{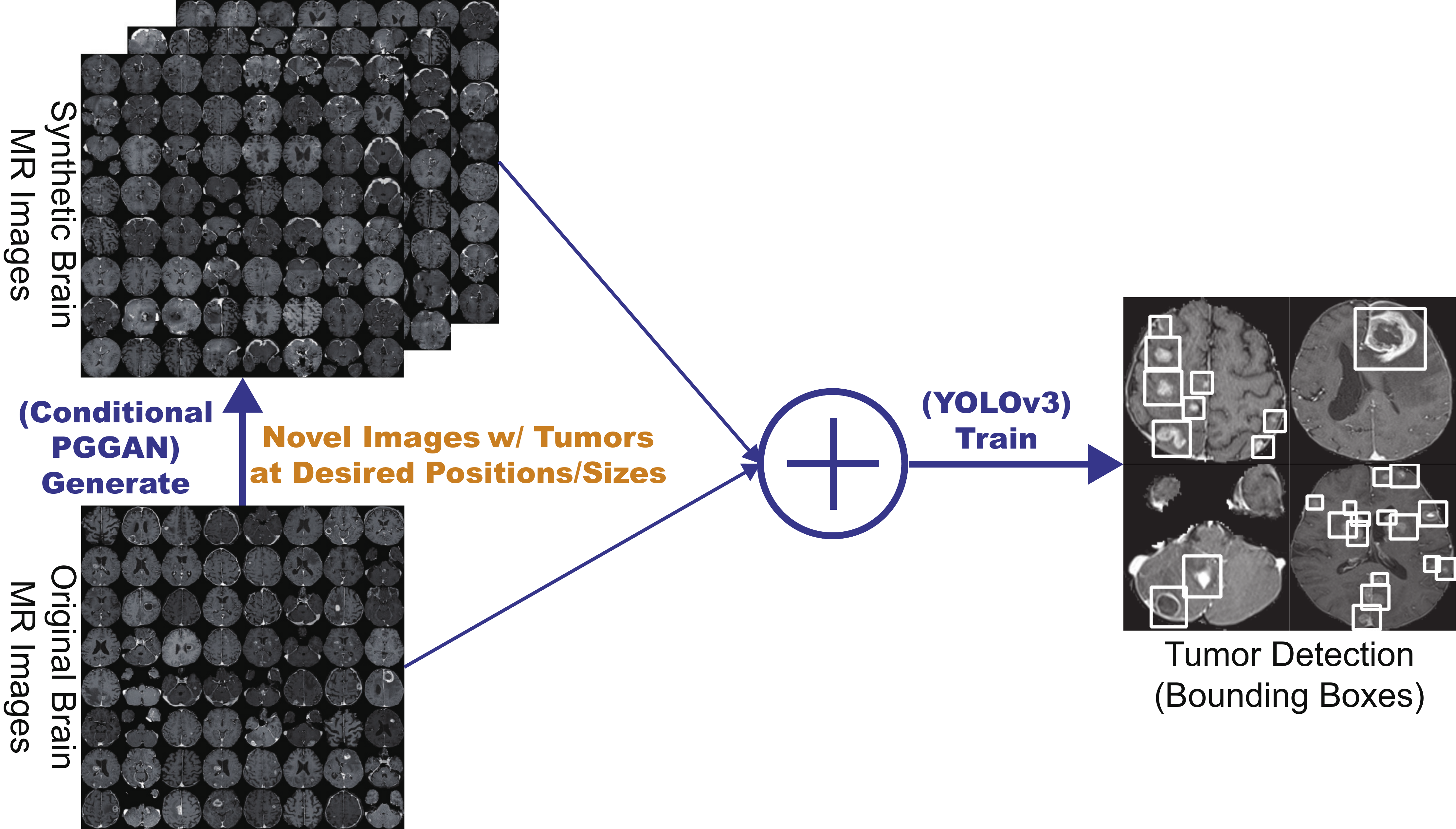}}
\caption{CPGGAN-based DA for better tumor detection: our CPGGANs generates a number of realistic/diverse brain MR images with tumors at desired positions/sizes based on bounding boxes, and the object detector uses them as additional training data.}
\label{fig1}
\end{figure}

So, how can we achieve high sensitivity in diagnosis using GANs with minimum annotation cost, based on highly-rough and inconsistent bounding boxes?
As an advanced data wrangling approach, we aim to generate GAN-based realistic and diverse $256\times256$ brain MR images with brain metastases at desired positions/sizes for accurate CNN-based tumor detection; this is clinically valuable for better diagnosis, prognosis, and treatment, since brain metastases are the most common intra-cranial tumors, getting prevalent as oncological therapies improve cancer patients' survival~\cite{arvold2016updates}. Conventional GANs cannot generate realistic $256 \times 256$ whole brain MR images conditioned on tumor positions/sizes under limited training data/highly-rough annotation~\cite{Han2}; since noise-to-image GANs cannot directly be conditioned on an image describing desired objects, we have to use image-to-image GANs (e.g., input both the conditioning image/random noise samples or the conditioning image alone with dropout noises~\cite{srivastava2014dropout} on a generator~\cite{isola2017image})---it results in unrealistic high-resolution MR images with odd artifacts due to the limited training data/rough annotation, tumor variations, and strong consistency in brain anatomy, unless we also input a benign image sacrificing image diversity.

Such a high-resolution whole image generation approach, not involving Regions of Interest (ROIs) alone, however, could facilitate detection because it provides more image details and most CNN architectures adopt around $256 \times 256$ input pixels. Therefore, as a conditional noise-to-image GAN not relying on an input benign image, we propose Conditional Progressive Growing of GANs (CPGGANs), incorporating highly-rough bounding box conditions incrementally into PGGANs~\cite{Karras} to naturally place tumors of random shape at desired positions/sizes on MR images. Moreover, we evaluate the generated images' realism \textit{via} Visual Turing Test~\cite{Salimans} by three expert physicians, and visualize the data distribution \textit{via} the t-Distributed Stochastic Neighbor Embedding (t-SNE) algorithm~\cite{Maaten}.
Using the synthetic images, our novel CPGGAN-based DA boosts $10\%$ sensitivity in diagnosis with clinically acceptable additional False Positives (FPs).
Surprisingly, we confirm that further realistic tumor appearance, judged by the physicians, does not contribute to detection performance.\\

\noindent \textbf{Research Questions.} We mainly address two questions:
\begin{itemize}
\item \textbf{PGGAN Conditioning:} How can we modify PGGANs to naturally place objects of random shape, unlike rigorous segmentation, at desired positions/sizes based on highly-rough bounding box masks?
\item \textbf{Medical Data Augmentation:} How can we balance the number of real and additional synthetic training data to achieve the best detection performance?\\
\end{itemize}

\noindent \textbf{Contributions.} Our main contributions are as follows:
\begin{itemize}
\item \textbf{Conditional Image Generation:} As the first bounding box-based $256 \times 256$ whole pathological image generation approach, CPGGANs can generate realistic/diverse images with objects naturally at desired positions/sizes; the generated images can play a vital role in clinical oncology applications, such as DA, data anonymization, and physician training.

\item \textbf{Misdiagnosis Prevention:} This study allows us to achieve high sensitivity in automatic CAD using small/fragmented medical imaging datasets with minimum annotation efforts based on highly-rough/inconsistent bounding boxes.

\item \textbf{Brain Metastases Detection:} This first bounding box-based brain metastases detection method successfully detects tumors exploiting CPGGAN-based DA.
\end{itemize}

\section{Generative Adversarial Networks}
In terms of realism and diversity, GANs~\cite{goodfellow} have shown great promise in image generation~\cite{ledig2017photo, karras2018style} through a two-player minimax game.
However, the two-player objective function triggers difficult training, accompanying artifacts and mode collapse~\cite{DBLP:journals/corr/GulrajaniAADC17} when generating high-resolution images, such as $256 \times 256$ ones~\cite{Radford}; to tackle this, multi-stage generative training methods have been proposed: AttnGAN uses attention-driven multi-stage refinement for fine-grained text-to-image generation~\cite{xu2018attngan}; PGGANs adopts incremental training procedures from low to high resolution for generating a realistic image~\cite{Karras}. Moreover, GAN-based $128 \times 128$ conditional image synthesis using a bounding box can control generated images' local properties~\cite{reed2016learning}. GANs can typically generate more realistic images than other common deep generative models, such as variational autoencoders~\cite{kingma2013auto} suffering from the injected noise and imperfect reconstruction due to a single objective function~\cite{mescheder2017adversarial}; thus, as a DA technique, most researchers chose GANs for facilitating classification~\cite{antoniou2017data, mariani2018bagan}, object detection~\cite{ouyang2018pedestrian,huang2018auggan}, and segmentation~\cite{zhu2018data} to tackle the lack of training data.

This GAN-based DA trend especially applies to medical imaging for handling various types of small/fragmented datasets from multiple scanners: researchers used noise-to-image GANs for improving classification on brain tumor/non-tumor MR~\cite{Han2} and liver lesion Computed Tomography (CT) images~\cite{frid2018gan}; others used image-to-image GANs focusing on ROI (i.e., small pathological areas) for improving segmentation on 3D brain tumor MR~\cite{shin2018medical} and 3D lung nodule CT images~\cite{jin2018ct}.

However, to the best of our knowledge, our work is the first GAN-based medical DA method using automatic bounding box annotation, despite 2D bounding boxes' cheap annotation cost compared with rigorous 3D segmentation. Moreover, unlike the ROI DA work generating only pedestrians without the background for pedestrian detection~\cite{ouyang2018pedestrian}, this is the first GAN-based whole image augmentation approach including the background, relying on bounding boxes, in computer vision.
Along with classic transformations of real images, a completely different approach---generating novel whole $256 \times 256$ brain MR images with tumors at desired positions/sizes using CPGGANs---may become a clinical breakthrough in terms of annotation cost.

\section{Materials and Methods}
\subsection{Brain Metastases Dataset}
As a new dataset for the first bounding box-based brain metastases detection, this paper uses a dataset of contrast-enhanced T1-weighted (T1c) brain axial MR images, collected by the authors (National Center for Global Health and Medicine, Tokyo, Japan) and currently not publicly available for ethical restrictions; for robust clinical applications, it contains $180$ brain metastatic cancer cases from multiple MRI scanners---those images differ in contrast, magnetic field strength (i.e., $1.5$ T, $3.0$ T), and matrix size (i.e., $190 \times 224$, $216 \times 256$, $256 \times 256$, $460 \times 460$ pixels). In the clinical practice, T1c MRI is well-established in brain metastases detection thanks to its high-contrast in the enhancing region. We also use additional brain MR images from $193$ normal subjects only for CPGGAN training, not in tumor detection, to confirm the effect of combining the normal and pathological images for training.

\begin{figure}[t]
  \centering
  \centerline{\includegraphics[width=\columnwidth]{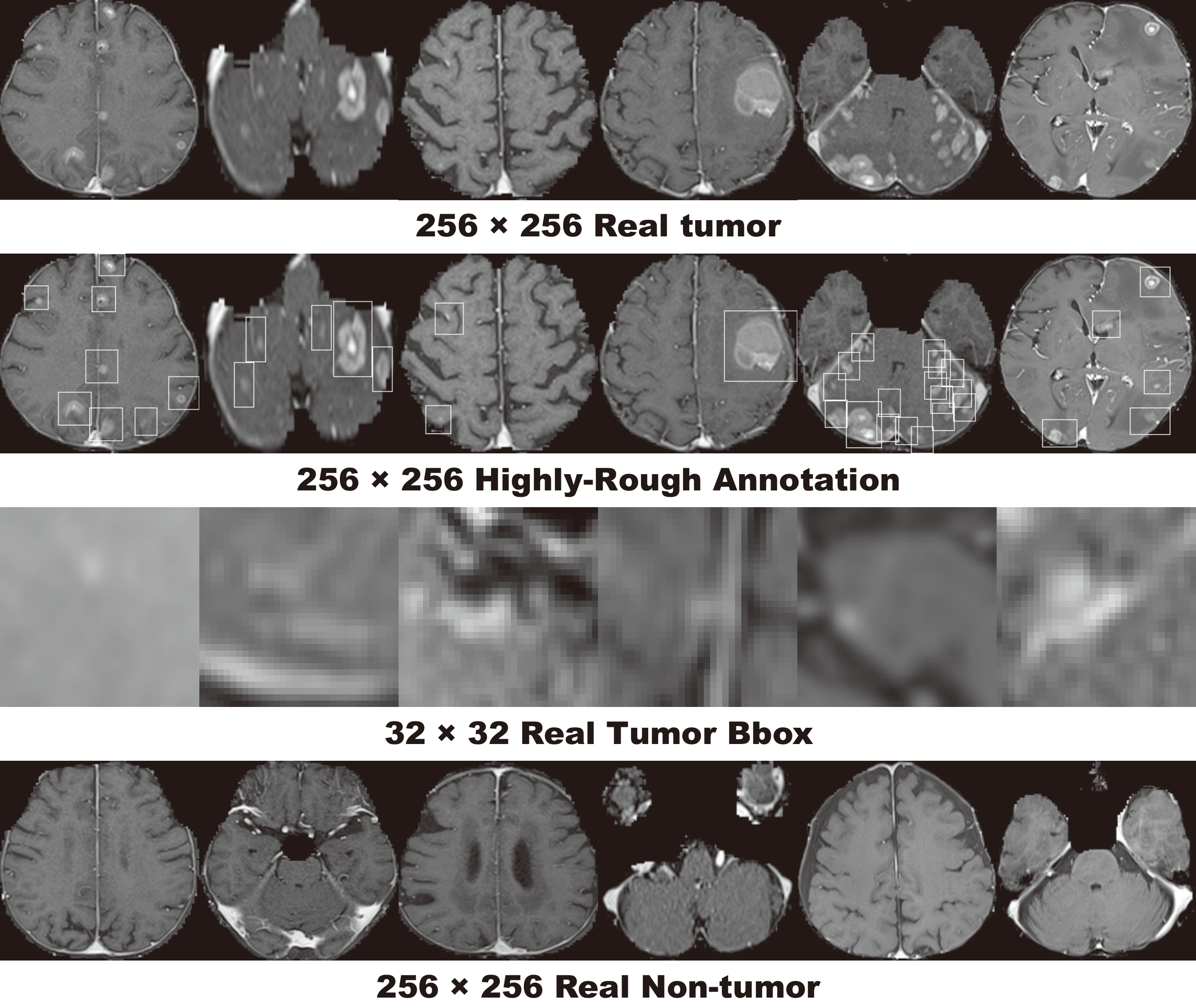}}
\caption{Example real $256 \times 256$ MR images with highly-rough annotation used for GAN training and resized $32 \times 32$ tumor bounding boxes.}
\label{fig2}
\end{figure}

\subsection{CPGGAN-based Image Generation}
\noindent \textbf{Data Preparation}
For tumor detection, our whole brain metastases dataset ($180$ patients) is divided into: (\textit{i}) a training set ($126$ patients); (\textit{ii}) a validation set ($18$ patients); (\textit{iii}) a test set ($36$ patients); only the training set is used for GAN training to be fair. Our experimental dataset consists of:
\begin{itemize}
\item Training set ($2,813$ images/$5,963$ bounding boxes);
\item Validation set ($337$ images/$616$ bounding boxes);
\item Test set ($947$ images/$3,094$ bounding boxes).
\end{itemize}


Our training set is relatively small/fragmented for CNN-based applications, considering that the same patient's tumor slices could convey very similar information. To confirm the effect of realism and diversity---provided by combining PGGANs and bounding box conditioning---on tumor detection, we compare the following GANs: (\textit{i}) CPGGANs trained only with the brain metastases images; (\textit{ii}) CPGGANs trained also with additional $16,962$ brain images from $193$ normal subjects; (\textit{iii}) Image-to-image GAN trained only with the brain metastases images. After skull-stripping on all images with various resolution, remaining brain parts are cropped and resized to $256 \times 256$ pixels (i.e., a power of $2$ for better GAN training). As Fig.~\ref{fig2} shows, we lazily annotate tumors with highly-rough and inconsistent bounding boxes to minimize expert physicians' labor.\\


\begin{figure}[t]
  \centering
  \centerline{\includegraphics[width=\columnwidth]{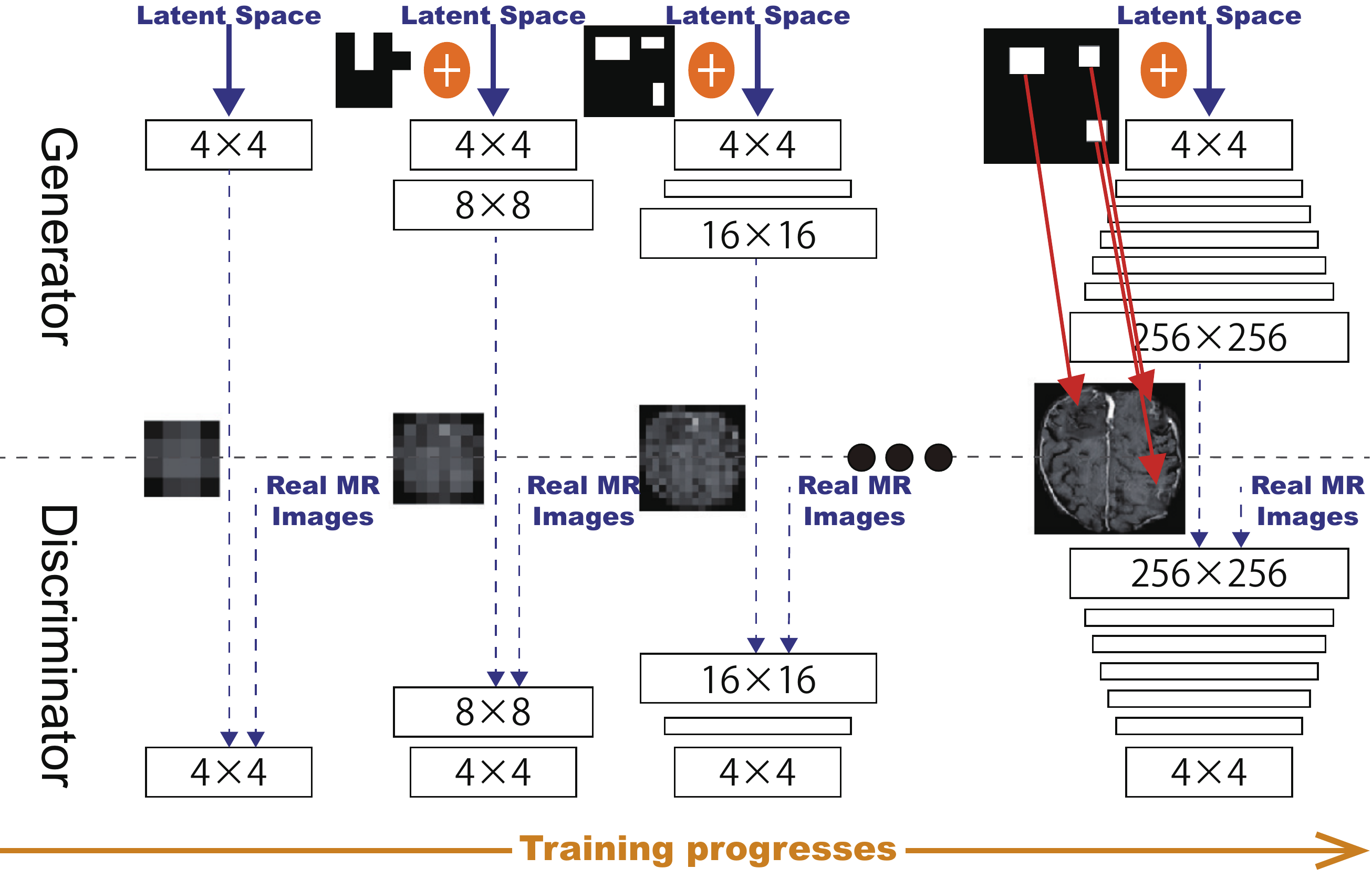}}
\caption{Proposed CPGGAN architecture for synthetic $256 \times 256$ MR image generation with tumors
at desired positions/sizes based on bounding boxes.}
\label{fig3}
\end{figure}

\noindent \textbf{CPGGANs} is a novel conditional noise-to-image training method for GANs, incorporating highly-rough bounding box conditions incrementally into PGGANs~\cite{Karras}, unlike conditional image-to-image GANs requiring rigorous segmentation masks~\cite{bailo2019red}. The original PGGANs exploits a progressively growing generator and discriminator: starting from low-resolution, newly-added layers model fine-grained details as training progresses. As Fig.~\ref{fig3} shows, we further condition the generator and discriminator to generate realistic and diverse $256 \times 256$ brain MR images with tumors of random shape at desired positions/sizes using only bounding boxes without an input benign image under limited training data/highly-rough annotation. Our modifications to the original PGGANs are as follows:

\begin{itemize}
\item Conditioning image: prepare a $256 \times 256$ black image (i.e., pixel value: $0$) with white bounding boxes (i.e., pixel value: $255$) describing tumor positions/sizes for attention;
\item Generator input: resize the conditioning image to the previous generator's output resolution/channel size and concatenate them (noise samples generate the first $4 \times 4$ images);
\item Discriminator input: concatenate the conditioning image with a real or synthetic image.
\end{itemize}

\noindent \textbf{CPGGAN Implementation Details}
We use the CPGGAN architecture with the Wasserstein loss using gradient penalty~\cite{DBLP:journals/corr/GulrajaniAADC17}:
\begin{eqnarray}\label{eq:wgan_gp}
\underset{{\tilde{y}\sim{\mathbb{P}_g}}}{\mathbb{E}}[D(\tilde{y})]-\underset{{y\sim{\mathbb{P}_r}}}{\mathbb{E}}[D(y)] +
\lambda\underset{{{\hat{y}}\sim{\mathbb{P}_{\hat{y}}}}} {\mathbb{E}}[(\left \| \nabla_{\hat{y}}{D({\hat{y}})} \right \|_2-1)^2]
\end{eqnarray}
where the discriminator $D$ belongs to the set of $1$-Lipschitz functions, $\mathbb{P}_r$ is the data distribution by the true data sample ${y}$, and $\mathbb{P}_g$ is the model distribution by the synthetic sample ${\tilde{y}}$ generated from the conditioning image noise samples using uniform distribution in $[-1, 1]$. The last term is gradient penalty for the random sample ${\hat{y}}\sim{\mathbb{P}_{\hat{y}}}$. 

Training lasts for $3,000,000$ steps with a batch size of $4$ and $2.0 \times 10^{-4}$ learning rate for the Adam optimizer \cite{kingma2014}. We flip the discriminator's real/synthetic labels once in three times for robustness. During testing, as tumor attention images, we use the annotation of training images with a random combination of horizontal/vertical flipping, width/height shift up to $10\%$, and zooming up to $10\%$; these CPGGAN-generated images are used as additional training images for tumor detection. \\

\noindent \textbf{Image-to-image GAN} is a conventional conditional GAN that generates brain MR images with tumors, concatenating a $256 \times 256$ conditioning image with noise samples for a generator input and concatenating the conditioning image with a real/synthetic image for a discriminator input, respectively. It uses a U-Net-like~\cite{ronneberger2015u} generator with $4$ convolutional/deconvolutional layers in encoders/decoders respectively with skip connections, along with a discriminator with $3$ decoders. We apply batch normalization~\cite{ioffe2015batch} to both convolution with LeakyReLU and deconvolution with ReLU. It follows the same implementation details as for the CPGGANs.

\subsection{YOLOv3-based Brain Metastases Detection}
\noindent \textbf{You Only Look Once v3 (YOLOv3)}~\cite{DBLP:journals/corr/abs-1804-02767} is a fast and accurate CNN-based object detector: unlike conventional classifier-based detectors, it divides the image into regions and predicts bounding boxes/probabilities for each region. We adopt YOLOv3 to detect brain metastases on MR images since its high efficiency can play a clinical role in real-time tumor alert; moreover, it shows very comparable results with $608 \times 608$ network resolution against other state-of-the-art detectors, such as Faster RCNN~\cite{ren2015faster}.

To confirm the effect of GAN-based DA, the following detection results are compared: (\textit{i}) $2,813$ real images without DA, (\textit{ii}), (\textit{iii}), (\textit{iv}) with  $4,000$/$8,000$/$12,000$ CPGGAN-based DA, (\textit{v}), (\textit{vi}), (\textit{vii}) with  $4,000$/$8,000$/$12,000$ CPGGAN-based DA, trained with additional normal brain images, (\textit{viii}), (\textit{ix}), (\textit{x}) with  $4,000$/$8,000$/$12,000$ image-to-image GAN-based DA. Due to the risk of overlooking the diagnosis $via$ medical imaging, higher sensitivity matters more than less FPs; thus, we aim to achieve higher sensitivity with a clinically acceptable number of FPs, adding the additional synthetic training images. Since our annotation is highly-rough, we calculate sensitivity/FPs per slice with both Intersection over Union (IoU) threshold 0.5 and 0.25. For better DA, GAN-generated images with unclear tumor appearance are manually discarded.\\

\noindent \textbf{YOLOv3 Implementation Details}
We use the YOLOv3 architecture with Darknet-53 as a backbone classifier and sum squared error between the predictions/ground truth as a loss:

\begin{align}
&\lambda_\text{coord} \sum_{i=0}^{S^2}\sum_{j=0}^B \mathbbm{1}_{ij}^\text{obj}\left[(x_i-\hat{x}_i)^2 + (y_i-\hat{y}_i)^2 \right] \nonumber\\&+ \lambda_\text{coord} \sum_{i=0}^{S^2}\sum_{j=0}^B \mathbbm{1}_{ij}^\text{obj}\left[\left(\sqrt{w_i}-\sqrt{\hat{w}_i}\right)^2 +\left(\sqrt{h_i}-\sqrt{\hat{h}_i}\right)^2 \right]\nonumber
\end{align}

\begin{align}
&+ \resizebox{.45\linewidth}{!}{$\sum_{i=0}^{S^2}\sum_{j=0}^B \mathbbm{1}_{ij}^\text{obj}(C_i - \hat{C}_i)^2$} + \resizebox{.45\linewidth}{!}{$\lambda_\text{noobj}\sum_{i=0}^{S^2}\sum_{j=0}^B \mathbbm{1}_{ij}^\text{noobj}(C_i - \hat{C}_i)^2 \nonumber$}\\
&+ \sum_{i=0}^{S^2} \mathbbm{1}_{i}^\text{obj}\sum_{c \in \text{classes}}(p_i(c) - \hat{p}_i(c))^2
\end{align}
where $x_i, y_i$ are the centroid location of an anchor box, $w_i, h_i$ are the width/height of the anchor, $C_i$ is the $\text{Objectness}$ (i.e., confidence score of whether an object exists), and $p_i(c)$ is the classification loss. Let $S^2$ and $B$ be the size of a feature map and  the number of anchor boxes, respectively. $\mathbbm{1}_{i}^\text{obj}$ is $1$ when an object exists in cell $i$ and otherwise $0$.

During training, we use a batch size of $64$ and $1.0 \times 10^{-3}$ learning rate for the Adam optimizer. The network resolution is set to $416 \times 416$ pixels during training and $608 \times 608$ pixels during validation/testing respectively to detect small tumors better. We recalculate the anchors at each DA setup. As classic DA, geometric/intensity transformations are also applied to both real/synthetic images during training to achieve the best performance. For testing, we pick the model with the best sensitivity on validation with detection threshold 0.1\%/IoU threshold 0.5 between $96,000$-$240,000$ steps to avoid severe FPs while achieving high sensitivity.

\begin{table*}[!t]
\caption{YOLOv3 brain metastases detection results with/without DA, using bounding boxes with detection threshold 0.1\%.}
\label{tab1}
\centering
\begin{tabular}{lrrrr}
\Hline\noalign{\smallskip}
 & \multicolumn{2}{c}{IoU $\geq$ 0.5}	& \multicolumn{2}{c}{IoU $\geq$ 0.25}\\
\bfseries  & {\bfseries Sensitivity} & \bfseries FPs per slice & {\bfseries  Sensitivity}  & \bfseries FPs per slice \\\noalign{\smallskip}\hline\noalign{\smallskip}
2,813 real images & 0.67 & 4.11 & 0.83 & 3.59\\
\noalign{\smallskip}\hline\noalign{\smallskip}
+ 4,000 CPGGAN-based DA & \textbf{0.77} & 7.64 & \textbf{0.91} & 7.18\\
+ 8,000 CPGGAN-based DA & 0.71 & 6.36 &0.87 & 5.85\\
+ 12,000 CPGGAN-based DA & 0.76 & 11.77 & \textbf{0.91} & 11.29\\
\noalign{\smallskip}\hline\noalign{\smallskip}
+ 4,000 CPGGAN-based DA (+ normal) & 0.69 & 7.16 & 0.86 & 6.60\\
+ 8,000 CPGGAN-based DA (+ normal) & 0.73 & 8.10 & 0.89 & 7.59\\
+ 12,000 CPGGAN-based DA (+ normal) & 0.74 & 9.42 & 0.89 & 8.95\\
\noalign{\smallskip}\hline\noalign{\smallskip}
+ 4,000 Image-to-Image GAN-based DA & 0.72 & 6.21 & 0.87 & 5.70\\
+ 8,000 Image-to-Image GAN-based DA & 0.68 & \textbf{3.50} & 0.84 & \textbf{2.99}\\
+ 12,000 Image-to-Image GAN-based DA & 0.74 & 7.20 & 0.89 & 6.72\\
\noalign{\smallskip}\Hline\noalign{\smallskip}
\end{tabular}
\end{table*}

\subsection{Clinical Validation \textit{via} Visual Turing Test}
To quantitatively evaluate how realistic the CPGGAN-based synthetic images are, we supply, in random order, to three expert physicians a random selection of $50$ real and $50$ synthetic brain metastases images. They take four tests in ascending order: (\textit{i}), (\textit{ii}) test 1, 2: real \textit{vs} CPGGAN-generated resized $32 \times 32$ tumor bounding boxes, trained without/with additional normal brain images; (\textit{iii}), (\textit{iv}) test 3, 4: real \textit{vs} CPGGAN-generated $256 \times 256$ MR images, trained without/with additional normal brain images.

Then, the physicians are asked to constantly classify them as real/synthetic, if needed, zooming/rotating them, without previous training stages revealing which is real/synthetic. Such Visual Turing Test~\cite{Salimans} can probe the human ability to identify attributes/relationships in images, also in evaluating GAN-generated images' appearance~\cite{Shrivastava}.
This similarly applies to medical images in a clinical environment, wherein physicians' specialty is critical~\cite{Han1,frid2018gan}.

\subsection{Visualization \textit{via} t-SNE}
To visually analyze the distribution of real/synthetic images, we use t-SNE~\cite{Maaten} on a random selection of:
\begin{itemize}
\item $500$ real tumor images;
\item $500$ CPGGAN-generated tumor images;
\item $500$ CPGGAN-generated tumor images, trained with additional normal brain images.
\end{itemize}
We normalize the input images to $[0, 1]$.

T-SNE is a machine learning algorithm for dimensionality reduction to represent high-dimensional data into a lower-dimensional (2D/3D) space. It non-linearly adapts to input data using perplexity to balance between the data's local and global aspects.\\

\noindent \textbf{t-SNE Implementation Details}
We use t-SNE with a perplexity of $100$ for $1,000$ iterations to get a 2D representation.

\begin{figure}[!t]
  \centering
  \centerline{\includegraphics[width=\columnwidth]{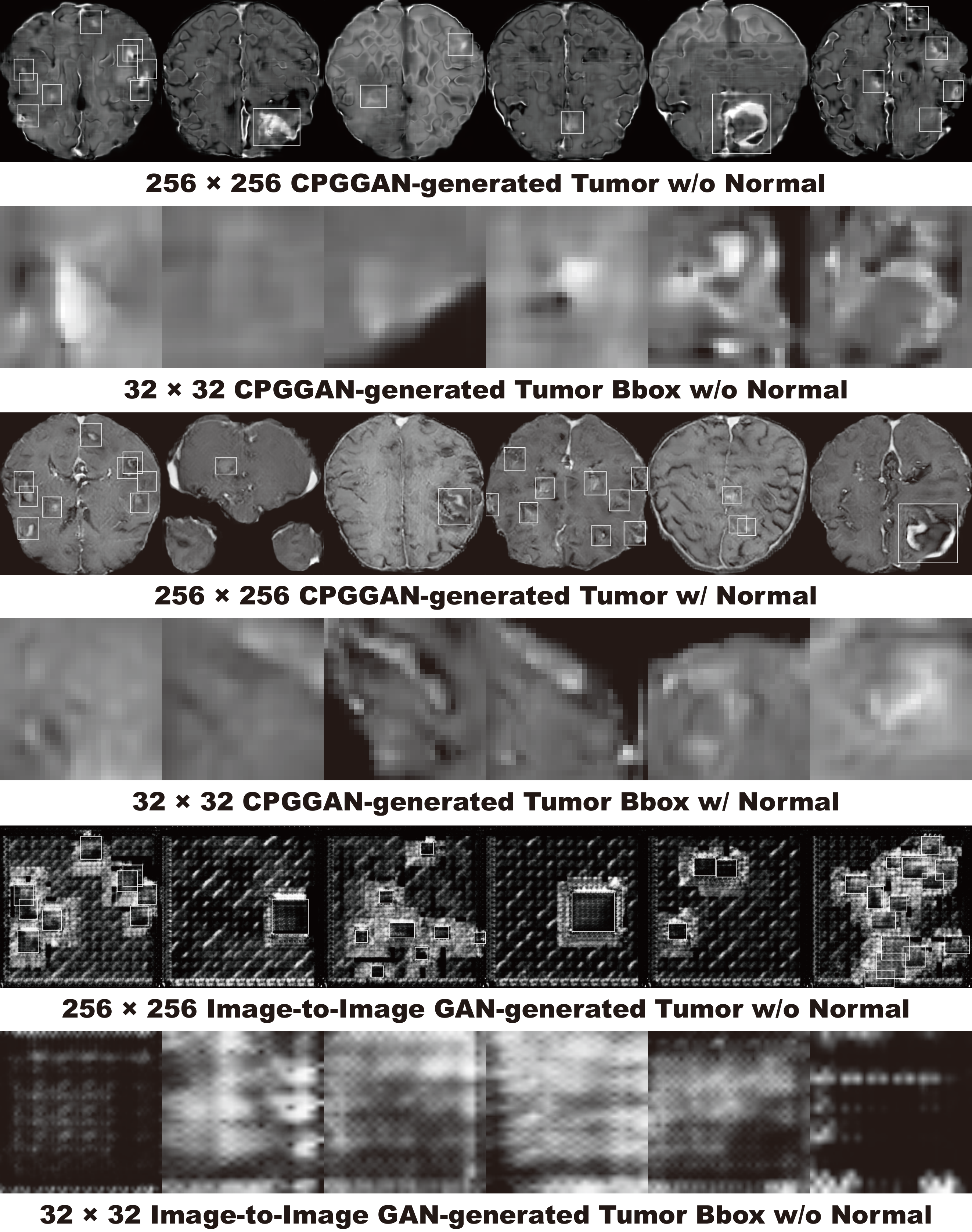}}
\caption{Example synthetic $256 \times 256$ MR images and resized $32 \times 32$ tumor bounding boxes yielded by (a), (b) CPGGANs trained without/with additional normal brain images; (c) image-to-image GAN trained without normal images.}
\label{fig4}
\end{figure}

\section{Results}
This section shows how CPGGANs and image-to-image GAN generate brain MR images. The results include instances of synthetic images and their influence on tumor detection, along with CPGGAN-generated images' evaluation \textit{via} Visual Turing Test and t-SNE. 
\subsection{MR Images Generated by CPGGANs}

Fig.~\ref{fig4} illustrates example GAN-generated images. CPGGANs successfully captures the T1c-specific texture and tumor appearance at desired positions/sizes. Since we use highly-rough bounding boxes, the synthetic tumor shape largely varies within the boxes. When trained with additional normal brain images, it clearly maintains the realism of the original images with less odd artifacts, including tumor bounding boxes, which the additional images do not include. However, as expected, image-to-image GAN, without progressive growing, generates clearly unrealistic images without an input benign image due to the limited training data/rough annotation.


\begin{figure*}[!t]
  \centering
  \centerline{\includegraphics[width=2\columnwidth]{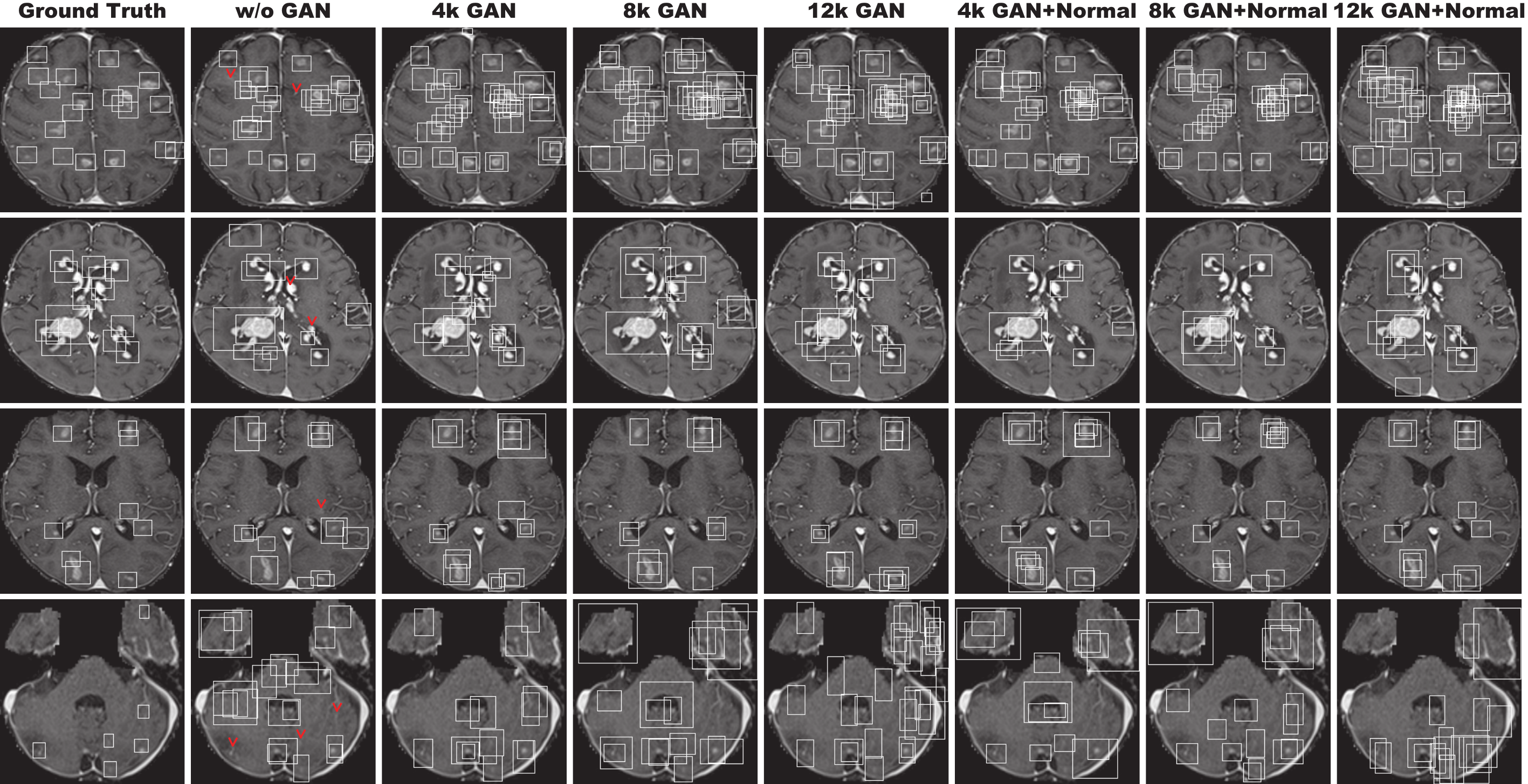}}
\caption{Example detection results obtained by the seven DA setups on four different images, compared against the ground truth: (a) ground truth; (b) without CPGGAN-based DA; (c), (d), (e) with $4\text{k}$/$8\text{k}$/$12\text{k}$ CPGGAN-based DA; (f), (g), (h) with $4\text{k}$/$8\text{k}$/$12\text{k}$ CPGGAN-based DA, trained with additional normal brain images. Red V symbols indicate the brain metastases undetected without CPGGAN-based DA, but detected with $4$k CPGGAN-based DA.}
\label{fig5}
\end{figure*}

\subsection{Brain Metastases Detection Results}
Table~\ref{tab1} shows the tumor detection results with/without GAN-based DA. As expected, the sensitivity remarkably increases with the additional synthetic training data while FPs per slice also increase. Adding more synthetic images generally leads to a higher amount of FPs, also detecting blood vessels that are small/hyper-intense on T1c MR images, very similarly to the enhanced tumor regions (i.e., the contrast agent perfuses throughout the blood vessels). However, surprisingly, adding only $4,000$ CPGGAN-generated images achieves the best sensitivity improvement by $0.10$ with IoU threshold $0.5$ and by $0.08$ with IoU threshold $0.25$, probably due to the real/synthetic training image balance---the improved training robustness achieves sensitivity $0.91$ with moderate IoU threshold $0.25$ despite our highly-rough bounding box annotation.

\begin{table*}[!t]
\caption{Visual Turing Test results by three physicians for classifying real \textit{vs} CPGGAN-generated images: (a), (b) Test 1, 2: resized $32 \times 32$ tumor bounding boxes, trained without/with additional normal brain images; (c), (d) Test 3, 4: $256 \times 256$ MR images, trained without/with normal brain images. Accuracy denotes the physicians' successful classification ratio between the real/synthetic images.}
\label{tab2}
\centering
\begin{tabular}{p{0.2em}lrrrrr}
\Hline\noalign{\smallskip}
& \bfseries  & \multicolumn{1}{c}{\bfseries Accuracy} & \bfseries Real Selected as Real & \bfseries Real as Synt & \bfseries Synt as Real & \bfseries Synt as Synt \\\noalign{\smallskip}\hline\noalign{\smallskip}
\parbox[t]{2mm}{\multirow{3}{*}{\rotatebox[origin=c]{270}{\textbf{\shortstack{\\Test 1}}}}} & Physician1 & 88\% & 40 & 10 & 2 & 48\\
& Physician2 & 95\% & 45 & 5 & 0 & 50\\
& Physician3 & 97\% & 49 & 1 & 2 & 48\\
\noalign{\smallskip}\hline\noalign{\smallskip}
\parbox[t]{2mm}{\multirow{3}{*}{\rotatebox[origin=c]{270}{\textbf{\shortstack{\\Test 2}}}}} & Physician1 & 81\% & 39 & 11 & 8 & 42\\
& Physician2 & 83\% & 43 & 7 & 10 & 40\\
& Physician3 & 91\% & 45 & 5 & 4 & 46\\
\noalign{\smallskip}\hline\noalign{\smallskip}
\parbox[t]{2mm}{\multirow{3}{*}{\rotatebox[origin=c]{270}{\textbf{\shortstack{\\Test 3}}}}} & Physician1 & 97\% & 47 & 3 & 0 & 50\\
& Physician2 & 96\% & 46 & 4 & 0 & 50\\
& Physician3 & 100\% & 50 & 0 & 0 & 50\\
\noalign{\smallskip}\hline\noalign{\smallskip}
\parbox[t]{2mm}{\multirow{3}{*}{\rotatebox[origin=c]{270}{\textbf{\shortstack{\\Test 4}}}}} & Physician1 & 91\% & 41 & 9 & 0 & 50\\
& Physician2 & 96\% & 48 & 2 & 2 & 48\\
& Physician3 & 100\% & 50 & 0 & 0 & 50\\
\noalign{\smallskip}\Hline\noalign{\smallskip}
\end{tabular}
\end{table*}

Fig.~\ref{fig5} also visually indicates that it can alleviate the risk of overlooking the tumor diagnosis with clinically acceptable FPs; in the clinical routine, the bounding boxes, highly-overlapping around tumors, only require a physician's single check by switching on/off transparent alpha-blended annotation on MR images. It should be noted that we cannot increase FPs to achieve such high sensitivity without CPGGAN-based DA. Moreover, our results reveal that further realism---associated with the additional normal brain images during training---does not contribute to detection performance, possibly as the training focuses less on tumor generation. Image-to-image GAN-based DA just moderately facilitates detection with less additional FPs, probably because the synthetic images have a distribution far from the real ones and thus their influence on detection is limited during testing.

\subsection{Visual Turing Test Results}

Table~\ref{tab2} shows the confusion matrix for the Visual Turing Test. The expert physicians easily recognize $256 \times 256$ synthetic images due to the lack of training data. However, when CPGGANs is trained with additional normal brain images, the experts classify a considerable number of synthetic tumor bounding boxes as real; it implies that the additional normal images remarkably facilitate the realism of both healthy and pathological brain parts while they do not include abnormality; thus, CPGGANs might perform as a tool to train medical students and radiology trainees when enough medical images are unavailable, such as abnormalities at rare positions/sizes. Such GAN applications are clinically prospective~\cite{finlayson2018towards}, considering the expert physicians' positive comments about the tumor realism.

\begin{figure}[t]
  \centering
  \centerline{\includegraphics[width=\columnwidth]{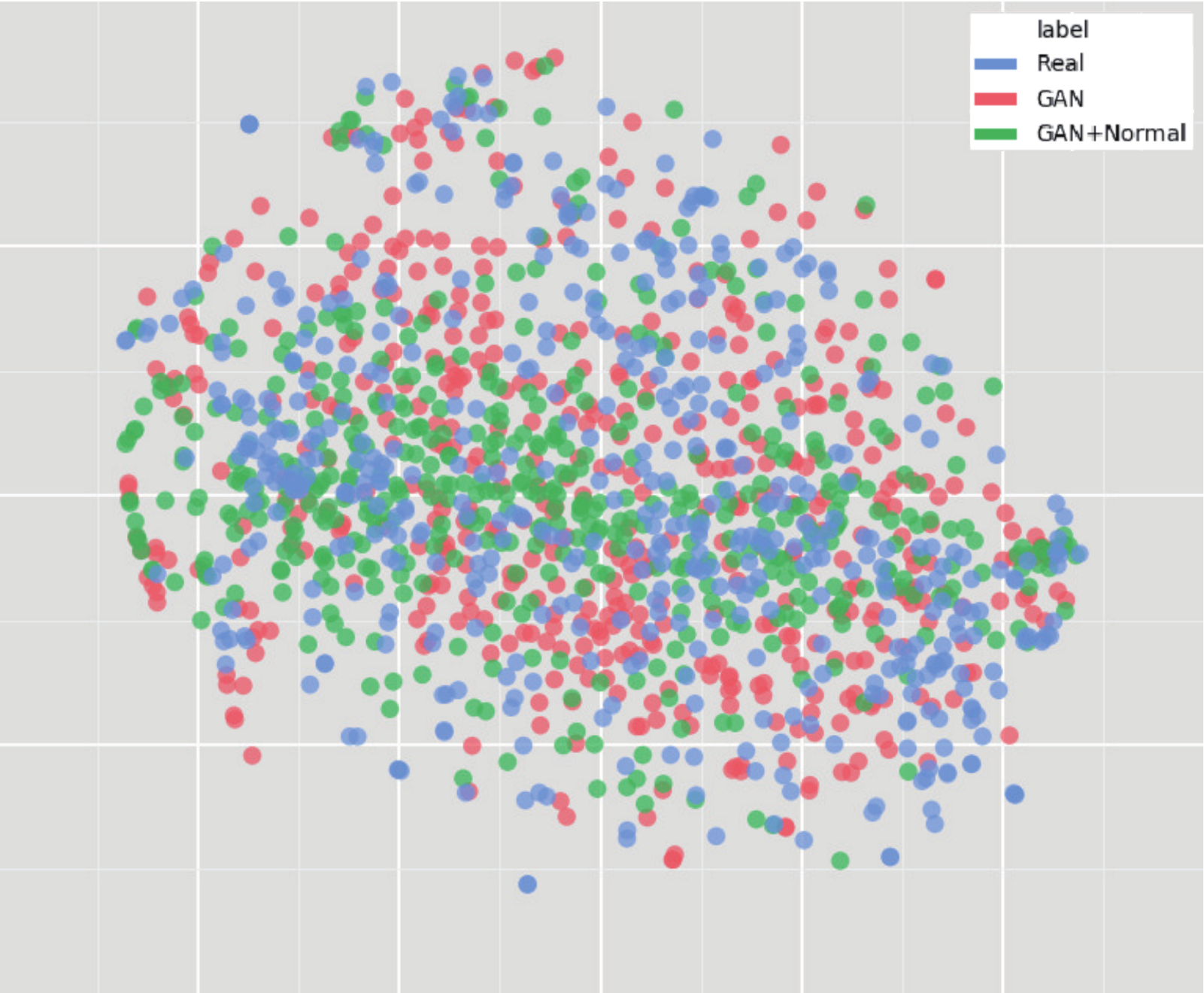}}
\caption{T-SNE results with $500$ $32 \times 32$ resized tumor bounding box images per each category: (a) Real tumor images; (b), (c) CPGGAN-generated tumor images, trained without/with additional normal brain images.}
\label{fig6}
\end{figure}

\begin{figure}[t]
  \centering
  \centerline{\includegraphics[width=\columnwidth]{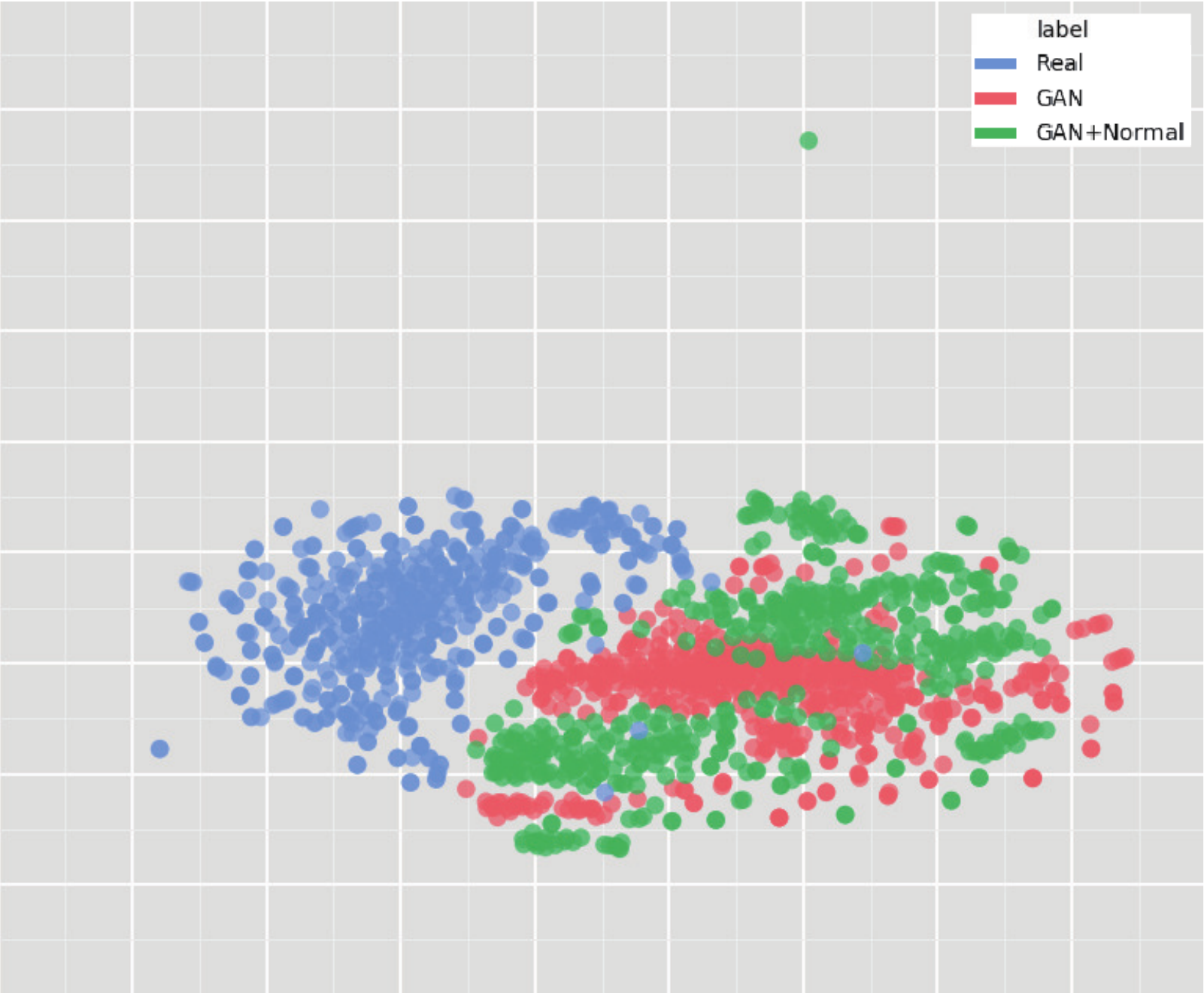}}
\caption{T-SNE results with $500$ $256 \times 256$ images per each category: (a) Real tumor images; (b), (c) CPGGAN-generated tumor images, trained without/with additional normal brain images.}
\label{fig7}
\end{figure}

\subsection{T-SNE Results}
As presented in Fig.~\ref{fig6}, synthetic tumor bounding boxes have a moderately similar distribution to real ones, but they also fill the real image distribution uncovered by the original dataset, implying their effective DA performance; especially, the CPGGAN-generated images trained without normal brain images distribute wider than the center-concentrating images trained with the normal brain images. Meanwhile, real/synthetic whole brain images clearly distribute differently, due to the real MR images' strong anatomical consistency (Fig.~\ref{fig7}). Considering the achieved high DA performance, the tumor (i.e., ROI) realism/diversity matter more than the whole image realism/diversity, since YOLOv3 look at an image patch instead of a whole image, similarly to most other CNN-based object detectors.

\section{Conclusion}
Without relying on an input benign image, our CPGGANs can generate realistic and diverse $256 \times 256$ MR images with brain metastases of random shape, unlike rigorous segmentation, naturally at desired positions/sizes, and achieve high sensitivity in tumor detection---even with small/fragmented training data from multiple MRI scanners and lazy annotation using highly-rough bounding boxes; in the context of intelligent data wrangling, this attributes to the CPGGANs' good generalization ability to incrementally synthesize conditional whole images with the real image distribution unfilled by the original dataset, improving the training robustness.

We confirm that the realism and diversity of the generated images, judged by three expert physicians $via$ Visual Turing Test, do not imply better detection performance; as the t-SNE results show, the CPGGAN-generated images, trained with additional non-tumor normal images, lack diversity probably because the training less focuses on tumors. Moreover, we notice that adding over-sufficient synthetic images leads to more FPs, but not always higher sensitivity, possibly due to the training data imbalance between real and synthetic images; as the t-SNE results reveal, the CPGGAN-generated tumor bonding boxes have a moderately similar---mutually complementary---distribution to the real ones; thus, GAN-overwhelming training images may decrease the necessary influence of the real samples and harm training, rather than providing robustness. Lastly, image-to-image GAN-based DA just moderately facilitates detection with less additional FPs, probably due to the lack of realism. However, further investigations are needed to maximize the effect of the CPGGAN-based medical image augmentation.


For example, we could verify the effect of further realism in return for less diversity by combining $\ell _1$ loss with the Wasserstein loss using gradient penalty for GAN training. We can also combine those CPGGAN-generated images, trained without/with additional brain images, similarly to ensemble learning~\cite{dietterich2002ensemble}. Lastly, we plan to define a new GAN loss function that directly optimizes the detection results, instead of realism, similarly to the three-player GAN for optimizing classification results~\cite{vandenhende2019three}.

Overall, minimizing expert physicians' annotation efforts, our novel CPGGAN-based DA approach sheds light on diagnostic and prognostic medical applications, not limited to brain metastases detection; future studies, especially on 3D bounding box detection with highly-rough annotation, are required to extend our promising results. Along with the DA, the CPGGANs has other potential clinical applications in oncology: (\textit{i}) A data anonymization tool to share patients' data outside their institution for training while preserving detection performance. Such a GAN-based application is reported in Shin~\textit{et al.}~\cite{shin2018medical}; (\textit{ii}) A physician training tool to display random synthetic medical images
with abnormalities at both common and rare positions/sizes, by training CPGGANs on highly unbalanced medical datasets (i.e., limited pathological and abundant normal samples,
respectively). It can help train medical students and radiology trainees despite infrastructural and legal constraints~\cite{finlayson2018towards}.

\section*{Acknowledgments}
This research was supported by AMED Grant Number JP18lk1010028.

\bibliographystyle{ACM-Reference-Format}
\bibliography{sample-base}


\begin{thebibliography}{38}


\ifx \showCODEN    \undefined \def \showCODEN     #1{\unskip}     \fi
\ifx \showDOI      \undefined \def \showDOI       #1{#1}\fi
\ifx \showISBNx    \undefined \def \showISBNx     #1{\unskip}     \fi
\ifx \showISBNxiii \undefined \def \showISBNxiii  #1{\unskip}     \fi
\ifx \showISSN     \undefined \def \showISSN      #1{\unskip}     \fi
\ifx \showLCCN     \undefined \def \showLCCN      #1{\unskip}     \fi
\ifx \shownote     \undefined \def \shownote      #1{#1}          \fi
\ifx \showarticletitle \undefined \def \showarticletitle #1{#1}   \fi
\ifx \showURL      \undefined \def \showURL       {\relax}        \fi
\providecommand\bibfield[2]{#2}
\providecommand\bibinfo[2]{#2}
\providecommand\natexlab[1]{#1}
\providecommand\showeprint[2][]{arXiv:#2}

\bibitem[\protect\citeauthoryear{Antoniou, Storkey, and Edwards}{Antoniou
  et~al\mbox{.}}{2017}]%
        {antoniou2017data}
\bibfield{author}{\bibinfo{person}{Antreas Antoniou}, \bibinfo{person}{Amos
  Storkey}, {and} \bibinfo{person}{Harrison Edwards}.}
  \bibinfo{year}{2017}\natexlab{}.
\newblock \showarticletitle{Data augmentation generative adversarial networks}.
\newblock \bibinfo{journal}{\emph{arXiv preprint arXiv:1711.04340}}
  (\bibinfo{year}{2017}).
\newblock


\bibitem[\protect\citeauthoryear{Arvold, Lee, Mehta, Margolin, Alexander,
  et~al\mbox{.}}{Arvold et~al\mbox{.}}{2016}]%
        {arvold2016updates}
\bibfield{author}{\bibinfo{person}{Nils~D Arvold}, \bibinfo{person}{Eudocia~Q
  Lee}, \bibinfo{person}{Minesh~P Mehta}, \bibinfo{person}{Kim Margolin},
  \bibinfo{person}{Brian~M Alexander}, {et~al\mbox{.}}}
  \bibinfo{year}{2016}\natexlab{}.
\newblock \showarticletitle{Updates in the management of brain metastases}.
\newblock \bibinfo{journal}{\emph{Neuro Oncol.}} \bibinfo{volume}{18},
  \bibinfo{number}{8} (\bibinfo{year}{2016}), \bibinfo{pages}{1043--1065}.
\newblock


\bibitem[\protect\citeauthoryear{Bailo, Ham, and Shin}{Bailo
  et~al\mbox{.}}{2019}]%
        {bailo2019red}
\bibfield{author}{\bibinfo{person}{Oleksandr Bailo}, \bibinfo{person}{DongShik
  Ham}, {and} \bibinfo{person}{Young~Min Shin}.}
  \bibinfo{year}{2019}\natexlab{}.
\newblock \showarticletitle{Red blood cell image generation for data
  augmentation using conditional generative adversarial networks}.
\newblock \bibinfo{journal}{\emph{arXiv preprint arXiv:1901.06219}}
  (\bibinfo{year}{2019}).
\newblock


\bibitem[\protect\citeauthoryear{Dietterich}{Dietterich}{2002}]%
        {dietterich2002ensemble}
\bibfield{author}{\bibinfo{person}{Thomas~G Dietterich}.}
  \bibinfo{year}{2002}\natexlab{}.
\newblock \showarticletitle{Ensemble learning}.
\newblock \bibinfo{journal}{\emph{The Handbook of Brain Theory and Neural
  Networks}}  \bibinfo{volume}{2} (\bibinfo{year}{2002}),
  \bibinfo{pages}{110--125}.
\newblock


\bibitem[\protect\citeauthoryear{Finlayson, Lee, Kohane, and
  Oakden-Rayner}{Finlayson et~al\mbox{.}}{2018}]%
        {finlayson2018towards}
\bibfield{author}{\bibinfo{person}{Samuel~G Finlayson},
  \bibinfo{person}{Hyunkwang Lee}, \bibinfo{person}{Isaac~S Kohane}, {and}
  \bibinfo{person}{Luke Oakden-Rayner}.} \bibinfo{year}{2018}\natexlab{}.
\newblock \showarticletitle{Towards generative adversarial networks as a new
  paradigm for radiology education}. In \bibinfo{booktitle}{\emph{Proc. Machine
  Learning for Health (ML4H) Workshop arXiv:1812.01547}}.
\newblock


\bibitem[\protect\citeauthoryear{Frid-Adar, Diamant, Klang, Amitai, Goldberger,
  and Greenspan}{Frid-Adar et~al\mbox{.}}{2018}]%
        {frid2018gan}
\bibfield{author}{\bibinfo{person}{Maayan Frid-Adar}, \bibinfo{person}{Idit
  Diamant}, \bibinfo{person}{Eyal Klang}, \bibinfo{person}{Michal Amitai},
  \bibinfo{person}{Jacob Goldberger}, {and} \bibinfo{person}{Hayit Greenspan}.}
  \bibinfo{year}{2018}\natexlab{}.
\newblock \showarticletitle{{GAN}-based synthetic medical image augmentation
  for increased {CNN} performance in liver lesion classification}.
\newblock \bibinfo{journal}{\emph{Neurocomputing}}  \bibinfo{volume}{321}
  (\bibinfo{year}{2018}), \bibinfo{pages}{321--331}.
\newblock


\bibitem[\protect\citeauthoryear{Goodfellow, Pouget-Abadie, Mirza, Xu,
  Warde-Farley, et~al\mbox{.}}{Goodfellow et~al\mbox{.}}{2014}]%
        {goodfellow}
\bibfield{author}{\bibinfo{person}{Ian Goodfellow}, \bibinfo{person}{Jean
  Pouget-Abadie}, \bibinfo{person}{Mehdi Mirza}, \bibinfo{person}{Bing Xu},
  \bibinfo{person}{David Warde-Farley}, {et~al\mbox{.}}}
  \bibinfo{year}{2014}\natexlab{}.
\newblock \showarticletitle{Generative adversarial nets}. In
  \bibinfo{booktitle}{\emph{Advances in Neural Information Processing Systems
  (NIPS)}}. \bibinfo{pages}{2672--2680}.
\newblock


\bibitem[\protect\citeauthoryear{Gulrajani, Ahmed, Arjovsky, Dumoulin, and
  Courville}{Gulrajani et~al\mbox{.}}{2017}]%
        {DBLP:journals/corr/GulrajaniAADC17}
\bibfield{author}{\bibinfo{person}{Ishaan Gulrajani}, \bibinfo{person}{Faruk
  Ahmed}, \bibinfo{person}{Mart{\'{\i}}n Arjovsky}, \bibinfo{person}{Vincent
  Dumoulin}, {and} \bibinfo{person}{Aaron~C. Courville}.}
  \bibinfo{year}{2017}\natexlab{}.
\newblock \showarticletitle{Improved Training of {Wasserstein GANs}}.
\newblock \bibinfo{journal}{\emph{arXiv preprint arXiv:1704.00028}}
  (\bibinfo{year}{2017}).
\newblock


\bibitem[\protect\citeauthoryear{Gulshan, Peng, Coram, Stumpe, Wu,
  et~al\mbox{.}}{Gulshan et~al\mbox{.}}{2016}]%
        {gulshan2016development}
\bibfield{author}{\bibinfo{person}{Varun Gulshan}, \bibinfo{person}{Lily Peng},
  \bibinfo{person}{Marc Coram}, \bibinfo{person}{Martin~C Stumpe},
  \bibinfo{person}{Derek Wu}, {et~al\mbox{.}}} \bibinfo{year}{2016}\natexlab{}.
\newblock \showarticletitle{Development and validation of a deep learning
  algorithm for detection of diabetic retinopathy in retinal fundus
  photographs}.
\newblock \bibinfo{journal}{\emph{JAMA}} \bibinfo{volume}{316},
  \bibinfo{number}{22} (\bibinfo{year}{2016}), \bibinfo{pages}{2402--2410}.
\newblock


\bibitem[\protect\citeauthoryear{Han, Hayashi, Rundo, Araki, Shimoda,
  et~al\mbox{.}}{Han et~al\mbox{.}}{2018}]%
        {Han1}
\bibfield{author}{\bibinfo{person}{Changhee Han}, \bibinfo{person}{Hideaki
  Hayashi}, \bibinfo{person}{Leonardo Rundo}, \bibinfo{person}{Ryosuke Araki},
  \bibinfo{person}{Wataru Shimoda}, {et~al\mbox{.}}}
  \bibinfo{year}{2018}\natexlab{}.
\newblock \showarticletitle{{GAN}-based synthetic brain {MR} image generation}.
  In \bibinfo{booktitle}{\emph{Proc. IEEE International Symposium on Biomedical
  Imaging (ISBI)}}. \bibinfo{pages}{734--738}.
\newblock


\bibitem[\protect\citeauthoryear{Han, Rundo, Araki, Furukawa, Mauri,
  et~al\mbox{.}}{Han et~al\mbox{.}}{2019}]%
        {Han2}
\bibfield{author}{\bibinfo{person}{Changhee Han}, \bibinfo{person}{Leonardo
  Rundo}, \bibinfo{person}{Ryosuke Araki}, \bibinfo{person}{Yujiro Furukawa},
  \bibinfo{person}{Giancarlo Mauri}, {et~al\mbox{.}}}
  \bibinfo{year}{2019}\natexlab{}.
\newblock \showarticletitle{Infinite brain {MR} images: {PGGAN}-based data
  augmentation for tumor detection}.
\newblock In \bibinfo{booktitle}{\emph{Neural Approaches to Dynamics of Signal
  Exchanges}}. \bibinfo{publisher}{Springer}.
\newblock
\urldef\tempurl%
\url{https://doi.org/10.1007/978-981-13-8950-4_27}
\showDOI{\tempurl}
\newblock
\shownote{(In press).}


\bibitem[\protect\citeauthoryear{Huang, Lin, Chen, Wu, Hsu, and Lai}{Huang
  et~al\mbox{.}}{2018}]%
        {huang2018auggan}
\bibfield{author}{\bibinfo{person}{Sheng-Wei Huang}, \bibinfo{person}{Che-Tsung
  Lin}, \bibinfo{person}{Shu-Ping Chen}, \bibinfo{person}{Yen-Yi Wu},
  \bibinfo{person}{Po-Hao Hsu}, {and} \bibinfo{person}{Shang-Hong Lai}.}
  \bibinfo{year}{2018}\natexlab{}.
\newblock \showarticletitle{{AugGAN}: cross domain adaptation with {GAN}-based
  data augmentation}. In \bibinfo{booktitle}{\emph{Proc. European Conference on
  Computer Vision (ECCV)}}. \bibinfo{pages}{718--731}.
\newblock


\bibitem[\protect\citeauthoryear{Ioffe and Szegedy}{Ioffe and Szegedy}{2015}]%
        {ioffe2015batch}
\bibfield{author}{\bibinfo{person}{Sergey Ioffe} {and}
  \bibinfo{person}{Christian Szegedy}.} \bibinfo{year}{2015}\natexlab{}.
\newblock \showarticletitle{Batch normalization: accelerating deep network
  training by reducing internal covariate shift}. In
  \bibinfo{booktitle}{\emph{Proc. International Conference on Learning
  Representations (ICLR) arXiv:1502.03167}}.
\newblock


\bibitem[\protect\citeauthoryear{Isola, Zhu, Zhou, and Efros}{Isola
  et~al\mbox{.}}{2017}]%
        {isola2017image}
\bibfield{author}{\bibinfo{person}{Phillip Isola}, \bibinfo{person}{Jun-Yan
  Zhu}, \bibinfo{person}{Tinghui Zhou}, {and} \bibinfo{person}{Alexei~A
  Efros}.} \bibinfo{year}{2017}\natexlab{}.
\newblock \showarticletitle{Image-to-image translation with conditional
  adversarial networks}. In \bibinfo{booktitle}{\emph{Proc. IEEE Conference on
  Computer Vision and Pattern Recognition (CVPR)}}.
  \bibinfo{pages}{1125--1134}.
\newblock


\bibitem[\protect\citeauthoryear{Jin, Xu, Tang, Harrison, and Mollura}{Jin
  et~al\mbox{.}}{2018}]%
        {jin2018ct}
\bibfield{author}{\bibinfo{person}{Dakai Jin}, \bibinfo{person}{Ziyue Xu},
  \bibinfo{person}{Youbao Tang}, \bibinfo{person}{Adam~P Harrison}, {and}
  \bibinfo{person}{Daniel~J Mollura}.} \bibinfo{year}{2018}\natexlab{}.
\newblock \showarticletitle{{CT}-realistic lung nodule simulation from {3D}
  conditional generative adversarial networks for robust lung segmentation}. In
  \bibinfo{booktitle}{\emph{Proc. International Conference on Medical Image
  Computing and Computer-Assisted Intervention (MICCAI)}}.
  \bibinfo{pages}{732--740}.
\newblock


\bibitem[\protect\citeauthoryear{Karras, Aila, Laine, and Lehtinen}{Karras
  et~al\mbox{.}}{2018a}]%
        {Karras}
\bibfield{author}{\bibinfo{person}{Tero Karras}, \bibinfo{person}{Timo Aila},
  \bibinfo{person}{Samuli Laine}, {and} \bibinfo{person}{Jaakko Lehtinen}.}
  \bibinfo{year}{2018}\natexlab{a}.
\newblock \showarticletitle{Progressive growing of {GAN}s for improved quality,
  stability, and variation}. In \bibinfo{booktitle}{\emph{Proc. International
  Conference on Learning Representations (ICLR) arXiv:1710.10196v3}}.
\newblock


\bibitem[\protect\citeauthoryear{Karras, Laine, and Aila}{Karras
  et~al\mbox{.}}{2018b}]%
        {karras2018style}
\bibfield{author}{\bibinfo{person}{Tero Karras}, \bibinfo{person}{Samuli
  Laine}, {and} \bibinfo{person}{Timo Aila}.} \bibinfo{year}{2018}\natexlab{b}.
\newblock \showarticletitle{A style-based generator architecture for generative
  adversarial networks}.
\newblock \bibinfo{journal}{\emph{arXiv preprint arXiv:1812.04948}}
  (\bibinfo{year}{2018}).
\newblock


\bibitem[\protect\citeauthoryear{Kingma and Ba}{Kingma and Ba}{2014}]%
        {kingma2014}
\bibfield{author}{\bibinfo{person}{Diederik~P Kingma} {and}
  \bibinfo{person}{Jimmy Ba}.} \bibinfo{year}{2014}\natexlab{}.
\newblock \showarticletitle{Adam: a method for stochastic optimization}.
\newblock \bibinfo{journal}{\emph{arXiv preprint arXiv:1412.6980}}
  (\bibinfo{year}{2014}).
\newblock


\bibitem[\protect\citeauthoryear{Kingma and Welling}{Kingma and
  Welling}{2013}]%
        {kingma2013auto}
\bibfield{author}{\bibinfo{person}{Diederik~P Kingma} {and}
  \bibinfo{person}{Max Welling}.} \bibinfo{year}{2013}\natexlab{}.
\newblock \showarticletitle{Auto-encoding variational bayes}. In
  \bibinfo{booktitle}{\emph{Proc. International Conference on Learning
  Representations (ICLR) arXiv:1312.6114}}.
\newblock


\bibitem[\protect\citeauthoryear{Ledig, Theis, Husz{\'a}r, Caballero,
  Cunningham, et~al\mbox{.}}{Ledig et~al\mbox{.}}{2017}]%
        {ledig2017photo}
\bibfield{author}{\bibinfo{person}{Christian Ledig}, \bibinfo{person}{Lucas
  Theis}, \bibinfo{person}{Ferenc Husz{\'a}r}, \bibinfo{person}{Jose
  Caballero}, \bibinfo{person}{Andrew Cunningham}, {et~al\mbox{.}}}
  \bibinfo{year}{2017}\natexlab{}.
\newblock \showarticletitle{Photo-realistic single image super-resolution using
  a generative adversarial network}. In \bibinfo{booktitle}{\emph{Proc. IEEE
  conference on Computer Vision and Pattern Recognition (CVPR)}}.
  \bibinfo{pages}{4681--4690}.
\newblock


\bibitem[\protect\citeauthoryear{Mariani, Scheidegger, Istrate, Bekas, and
  Malossi}{Mariani et~al\mbox{.}}{2018}]%
        {mariani2018bagan}
\bibfield{author}{\bibinfo{person}{Giovanni Mariani}, \bibinfo{person}{Florian
  Scheidegger}, \bibinfo{person}{Roxana Istrate}, \bibinfo{person}{Costas
  Bekas}, {and} \bibinfo{person}{Cristiano Malossi}.}
  \bibinfo{year}{2018}\natexlab{}.
\newblock \showarticletitle{{BAGAN}: data augmentation with balancing {GAN}}.
\newblock \bibinfo{journal}{\emph{arXiv preprint arXiv:1803.09655}}
  (\bibinfo{year}{2018}).
\newblock


\bibitem[\protect\citeauthoryear{Mescheder, Nowozin, and Geiger}{Mescheder
  et~al\mbox{.}}{2017}]%
        {mescheder2017adversarial}
\bibfield{author}{\bibinfo{person}{Lars Mescheder}, \bibinfo{person}{Sebastian
  Nowozin}, {and} \bibinfo{person}{Andreas Geiger}.}
  \bibinfo{year}{2017}\natexlab{}.
\newblock \showarticletitle{Adversarial variational bayes: unifying variational
  autoencoders and generative adversarial networks}. In
  \bibinfo{booktitle}{\emph{Proc. International Conference on Machine Learning
  (ICML)}}. \bibinfo{pages}{2391--2400}.
\newblock


\bibitem[\protect\citeauthoryear{Milletari, Navab, and Ahmadi}{Milletari
  et~al\mbox{.}}{2016}]%
        {milletari2016v}
\bibfield{author}{\bibinfo{person}{Fausto Milletari}, \bibinfo{person}{Nassir
  Navab}, {and} \bibinfo{person}{Seyed-Ahmad Ahmadi}.}
  \bibinfo{year}{2016}\natexlab{}.
\newblock \showarticletitle{{V-Net}: fully convolutional neural networks for
  volumetric medical image segmentation}. In \bibinfo{booktitle}{\emph{Proc.
  IEEE International Conference on 3D Vision (3DV)}}.
  \bibinfo{pages}{565--571}.
\newblock


\bibitem[\protect\citeauthoryear{Ouyang, Cheng, Jiang, Li, and Zhou}{Ouyang
  et~al\mbox{.}}{2018}]%
        {ouyang2018pedestrian}
\bibfield{author}{\bibinfo{person}{Xi Ouyang}, \bibinfo{person}{Yu Cheng},
  \bibinfo{person}{Yifan Jiang}, \bibinfo{person}{Chun-Liang Li}, {and}
  \bibinfo{person}{Pan Zhou}.} \bibinfo{year}{2018}\natexlab{}.
\newblock \showarticletitle{Pedestrian-Synthesis-{GAN}: generating pedestrian
  data in real scene and beyond}.
\newblock \bibinfo{journal}{\emph{arXiv preprint arXiv: 1804.02047}}
  (\bibinfo{year}{2018}).
\newblock


\bibitem[\protect\citeauthoryear{Radford, Metz, and Chintala}{Radford
  et~al\mbox{.}}{2016}]%
        {Radford}
\bibfield{author}{\bibinfo{person}{Alec Radford}, \bibinfo{person}{Luke Metz},
  {and} \bibinfo{person}{Soumith Chintala}.} \bibinfo{year}{2016}\natexlab{}.
\newblock \showarticletitle{Unsupervised representation learning with deep
  convolutional generative adversarial networks}. In
  \bibinfo{booktitle}{\emph{Proc. International Conference on Learning
  Representations (ICLR) arXiv:1511.06434}}.
\newblock


\bibitem[\protect\citeauthoryear{Redmon and Farhadi}{Redmon and
  Farhadi}{2018}]%
        {DBLP:journals/corr/abs-1804-02767}
\bibfield{author}{\bibinfo{person}{Joseph Redmon} {and} \bibinfo{person}{Ali
  Farhadi}.} \bibinfo{year}{2018}\natexlab{}.
\newblock \showarticletitle{{YOLOv3}: an incremental improvement}.
\newblock \bibinfo{journal}{\emph{arXiv preprint arXiv:1804.02767}}
  (\bibinfo{year}{2018}).
\newblock


\bibitem[\protect\citeauthoryear{Reed, Akata, Mohan, Tenka, Schiele, and
  Lee}{Reed et~al\mbox{.}}{2016}]%
        {reed2016learning}
\bibfield{author}{\bibinfo{person}{Scott~E Reed}, \bibinfo{person}{Zeynep
  Akata}, \bibinfo{person}{Santosh Mohan}, \bibinfo{person}{Samuel Tenka},
  \bibinfo{person}{Bernt Schiele}, {and} \bibinfo{person}{Honglak Lee}.}
  \bibinfo{year}{2016}\natexlab{}.
\newblock \showarticletitle{Learning what and where to draw}. In
  \bibinfo{booktitle}{\emph{Advances in Neural Information Processing Systems
  (NIPS)}}. \bibinfo{pages}{217--225}.
\newblock


\bibitem[\protect\citeauthoryear{Ren, He, Girshick, and Sun}{Ren
  et~al\mbox{.}}{2015}]%
        {ren2015faster}
\bibfield{author}{\bibinfo{person}{Shaoqing Ren}, \bibinfo{person}{Kaiming He},
  \bibinfo{person}{Ross Girshick}, {and} \bibinfo{person}{Jian Sun}.}
  \bibinfo{year}{2015}\natexlab{}.
\newblock \showarticletitle{Faster {R-CNN}: towards real-time object detection
  with region proposal networks}. In \bibinfo{booktitle}{\emph{Advances in
  Neural Information Processing Systems (NIPS)}}. \bibinfo{pages}{91--99}.
\newblock


\bibitem[\protect\citeauthoryear{Ronneberger, Fischer, and Brox}{Ronneberger
  et~al\mbox{.}}{2015}]%
        {ronneberger2015u}
\bibfield{author}{\bibinfo{person}{Olaf Ronneberger}, \bibinfo{person}{Philipp
  Fischer}, {and} \bibinfo{person}{Thomas Brox}.}
  \bibinfo{year}{2015}\natexlab{}.
\newblock \showarticletitle{{U-Net}: convolutional networks for biomedical
  image segmentation}. In \bibinfo{booktitle}{\emph{Proc. International
  Conference on Medical Image Computing and Computer-Assisted Intervention
  (MICCAI)}}. \bibinfo{pages}{234--241}.
\newblock


\bibitem[\protect\citeauthoryear{Rundo, Han, Nagano, Zhang, Hataya,
  et~al\mbox{.}}{Rundo et~al\mbox{.}}{2019}]%
        {Rundo}
\bibfield{author}{\bibinfo{person}{Leonardo Rundo}, \bibinfo{person}{Changhee
  Han}, \bibinfo{person}{Yudai Nagano}, \bibinfo{person}{Jin Zhang},
  \bibinfo{person}{Ryuichiro Hataya}, {et~al\mbox{.}}}
  \bibinfo{year}{2019}\natexlab{}.
\newblock \showarticletitle{{USE-Net}: incorporating squeeze-and-excitation
  blocks into {U-Net} for prostate zonal segmentation of multi-institutional
  {MRI} datasets}.
\newblock \bibinfo{journal}{\emph{Neurocomputing}} (\bibinfo{year}{2019}).
\newblock
\newblock
\shownote{(In press).}


\bibitem[\protect\citeauthoryear{Salimans, Goodfellow, Zaremba, Cheung,
  Radford, and Chen}{Salimans et~al\mbox{.}}{2016}]%
        {Salimans}
\bibfield{author}{\bibinfo{person}{Tim Salimans}, \bibinfo{person}{Ian
  Goodfellow}, \bibinfo{person}{Wojciech Zaremba}, \bibinfo{person}{Vicki
  Cheung}, \bibinfo{person}{Alec Radford}, {and} \bibinfo{person}{Xi Chen}.}
  \bibinfo{year}{2016}\natexlab{}.
\newblock \showarticletitle{Improved techniques for training {GAN}s}. In
  \bibinfo{booktitle}{\emph{Advances in Neural Information Processing Systems
  (NIPS)}}. \bibinfo{pages}{2234--2242}.
\newblock


\bibitem[\protect\citeauthoryear{Shin, Tenenholtz, Rogers, Schwarz, Senjem,
  et~al\mbox{.}}{Shin et~al\mbox{.}}{2018}]%
        {shin2018medical}
\bibfield{author}{\bibinfo{person}{Hoo-Chang Shin}, \bibinfo{person}{Neil~A
  Tenenholtz}, \bibinfo{person}{Jameson~K Rogers},
  \bibinfo{person}{Christopher~G Schwarz}, \bibinfo{person}{Matthew~L Senjem},
  {et~al\mbox{.}}} \bibinfo{year}{2018}\natexlab{}.
\newblock \showarticletitle{Medical image synthesis for data augmentation and
  anonymization using generative adversarial networks}. In
  \bibinfo{booktitle}{\emph{International Workshop on Simulation and Synthesis
  in Medical Imaging (SASHIMI)}}. \bibinfo{pages}{1--11}.
\newblock


\bibitem[\protect\citeauthoryear{Shrivastava, Pfister, Tuzel, Susskind, Wang,
  and Webb}{Shrivastava et~al\mbox{.}}{2017}]%
        {Shrivastava}
\bibfield{author}{\bibinfo{person}{Ashish Shrivastava}, \bibinfo{person}{Tomas
  Pfister}, \bibinfo{person}{Oncel Tuzel}, \bibinfo{person}{Joshua Susskind},
  \bibinfo{person}{Wenda Wang}, {and} \bibinfo{person}{Russell Webb}.}
  \bibinfo{year}{2017}\natexlab{}.
\newblock \showarticletitle{Learning from simulated and unsupervised images
  through adversarial training}. In \bibinfo{booktitle}{\emph{Proc. IEEE
  Conference on Computer Vision and Pattern Recognition (CVPR)}}.
  \bibinfo{pages}{2107--2116}.
\newblock


\bibitem[\protect\citeauthoryear{Srivastava, Hinton, Krizhevsky, Sutskever, and
  Salakhutdinov}{Srivastava et~al\mbox{.}}{2014}]%
        {srivastava2014dropout}
\bibfield{author}{\bibinfo{person}{Nitish Srivastava},
  \bibinfo{person}{Geoffrey Hinton}, \bibinfo{person}{Alex Krizhevsky},
  \bibinfo{person}{Ilya Sutskever}, {and} \bibinfo{person}{Ruslan
  Salakhutdinov}.} \bibinfo{year}{2014}\natexlab{}.
\newblock \showarticletitle{Dropout: a simple way to prevent neural networks
  from overfitting}.
\newblock \bibinfo{journal}{\emph{J. Mach. Learn. Res.}} \bibinfo{volume}{15},
  \bibinfo{number}{1} (\bibinfo{year}{2014}), \bibinfo{pages}{1929--1958}.
\newblock


\bibitem[\protect\citeauthoryear{van~der Maaten and Hinton}{van~der Maaten and
  Hinton}{2008}]%
        {Maaten}
\bibfield{author}{\bibinfo{person}{Laurens van~der Maaten} {and}
  \bibinfo{person}{Geoffrey Hinton}.} \bibinfo{year}{2008}\natexlab{}.
\newblock \showarticletitle{Visualizing data using {t-SNE}}.
\newblock \bibinfo{journal}{\emph{J. Mach. Learn. Res.}}  \bibinfo{volume}{9}
  (\bibinfo{year}{2008}), \bibinfo{pages}{2579--2605}.
\newblock


\bibitem[\protect\citeauthoryear{Vandenhende, De~Brabandere, Neven, and
  Van~Gool}{Vandenhende et~al\mbox{.}}{2019}]%
        {vandenhende2019three}
\bibfield{author}{\bibinfo{person}{Simon Vandenhende}, \bibinfo{person}{Bert
  De~Brabandere}, \bibinfo{person}{Davy Neven}, {and} \bibinfo{person}{Luc
  Van~Gool}.} \bibinfo{year}{2019}\natexlab{}.
\newblock \showarticletitle{A three-player {GAN}: generating hard samples to
  improve classification networks}. In \bibinfo{booktitle}{\emph{Proc.
  International Conference on Machine Vision Applications (MVA)
  arXiv:1903.03496}}.
\newblock


\bibitem[\protect\citeauthoryear{Xu, Zhang, Huang, Zhang, Gan,
  et~al\mbox{.}}{Xu et~al\mbox{.}}{2018}]%
        {xu2018attngan}
\bibfield{author}{\bibinfo{person}{Tao Xu}, \bibinfo{person}{Pengchuan Zhang},
  \bibinfo{person}{Qiuyuan Huang}, \bibinfo{person}{Han Zhang},
  \bibinfo{person}{Zhe Gan}, {et~al\mbox{.}}} \bibinfo{year}{2018}\natexlab{}.
\newblock \showarticletitle{{AttnGAN}: fine-grained text to image generation
  with attentional generative adversarial networks}. In
  \bibinfo{booktitle}{\emph{Proc. IEEE Conference on Computer Vision and
  Pattern Recognition (CVPR)}}. \bibinfo{pages}{1316--1324}.
\newblock


\bibitem[\protect\citeauthoryear{Zhu, Aoun, Krijn, Vanschoren, and Campus}{Zhu
  et~al\mbox{.}}{2018}]%
        {zhu2018data}
\bibfield{author}{\bibinfo{person}{Yezi Zhu}, \bibinfo{person}{Marc Aoun},
  \bibinfo{person}{Marcel Krijn}, \bibinfo{person}{Joaquin Vanschoren}, {and}
  \bibinfo{person}{High~Tech Campus}.} \bibinfo{year}{2018}\natexlab{}.
\newblock \showarticletitle{Data augmentation using conditional generative
  adversarial networks for leaf counting in arabidopsis plants}. In
  \bibinfo{booktitle}{\emph{Proc. Computer Vision Problems in Plant Phenotyping
  (CVPPP)}}. \bibinfo{pages}{324--334}.
\newblock


\end{thebibliography}

\end{document}